\documentclass[letterpaper, 10 pt, conference]{ieeeconf}  % Comment this line out
\IEEEoverridecommandlockouts                              % This command is only
\usepackage{booktabs}% For formal tables
\usepackage{todonotes}
\usepackage{subfigure}
\usepackage[noadjust]{cite}
\usepackage{algorithm}
\usepackage{amsmath}
\usepackage{algorithmic}
\usepackage{mcode}
\usepackage[hidelinks]{hyperref}
\usepackage{siunitx}
\usepackage{caption}
\usepackage{tabu}
\usepackage{multirow}
\usepackage{graphicx}
\title{\LARGE \bf Gaze-based, Context-aware Robotic System for Assisted Reaching and Grasping*}

\author{Ali Shafti\textsuperscript{1}, Pavel Orlov\textsuperscript{1} and A. Aldo Faisal
\thanks{*Research supported by eNHANCE (\href{http://www.enhance-motion.eu}{http://www.enhance-motion.eu}) under the European Union's Horizon 2020 research and innovation programme grant agreement No. 644000.}
\thanks{\textsuperscript{1}A. Shafti and P. Orlov contributed equally to this work. Along with A. A. Faisal, they are with the Brain and Behaviour Lab, Dept. of Computing and Dept. of Bioengineering, Imperial College London, SW7 2AZ, London, UK.
        {\tt\small a.shafti, p.orlov, a.faisal@imperial.ac.uk}}
}

\begin{document}

\maketitle
\thispagestyle{empty}
\pagestyle{empty}

%%%%%%%%%%%%%%%%%%%%%%%%%%%%%%%%%%%%%%%%%%%%%%%%%%%%%%%%%%%%%%%%%%%%%%%%%%%%%%%%
\begin{abstract}
Assistive robotic systems endeavour to support those with movement disabilities, enabling them to move again and regain functionality. Main issue with these systems is the complexity of their low-level control, and how to translate this to simpler, higher level commands that are easy and intuitive for a human user to interact with. We have created a multi-modal system, consisting of different sensing, decision making and actuating modalities, leading to intuitive, human-in-the-loop assistive robotics. The system takes its cue from the user's gaze, to decode their intentions and implement low-level motion actions to achieve high-level tasks. This results in the user simply having to look at the objects of interest, for the robotic system to assist them in reaching for those objects, grasping them, and using them to interact with other objects. We present our method for 3D gaze estimation, and grammars-based implementation of sequences of action with the robotic system. The 3D gaze estimation is evaluated with 8 subjects, showing an overall accuracy of 4.68\textpm0.14cm. The full system is tested with 5 subjects, showing successful implementation of 100\% of reach to gaze point actions and full implementation of pick and place tasks in 96\%, and pick and pour tasks in 76\% of cases. Finally we present a discussion on our results and what future work is needed to improve the system.
\end{abstract}
%%%%%%%%%%%%%%%%%%%%%%%%%%%%%%%%%%%%%%%%%%%%%%%%%%%%%%%%%%%%%%%%%%%%%%%%%%%%%%%%
\section{Introduction}
Limitations in human upper limb movements can be a result of spinal cord injuries, neurodegenerative diseases or strokes. These adversely affect a person's ability for basic activities of daily life. Robotic solutions are being devised as alternative actuators, to assist with these issues. Devices are presented in the form of exoskeletons \cite{rocon2007design,bortole2015h2}, prosthetics \cite{cipriani2008shared}, orthotics \cite{sanchez2006automating} and supernumerary robotic limbs \cite{wu2015hold,hussain2016soft,ciullo2018analytical,cunningham2018supernumerary} which can replace or augment actuation. For such systems, the user's control interface is typically either residual motion (e.g. sip and puff \cite{cunningham2014jamster}) or neural interfaces (e.g. muscle activity \cite{englehart2003robust} or brain-computer interfaces \cite{ajiboye2017restoration}). These interfaces are not available to all patients, and/or require invasive procedures and long training times for the user to be adept with their use \cite{ajiboye2017restoration,muelling2017autonomy,downey2018intracortical}. The degrees of freedom tend to exceed the available number of independent channels within the above interfaces, and therefore result either in simplified device capabilities or difficulties in user control.
\begin{figure}[tp]
\includegraphics[width=\columnwidth]{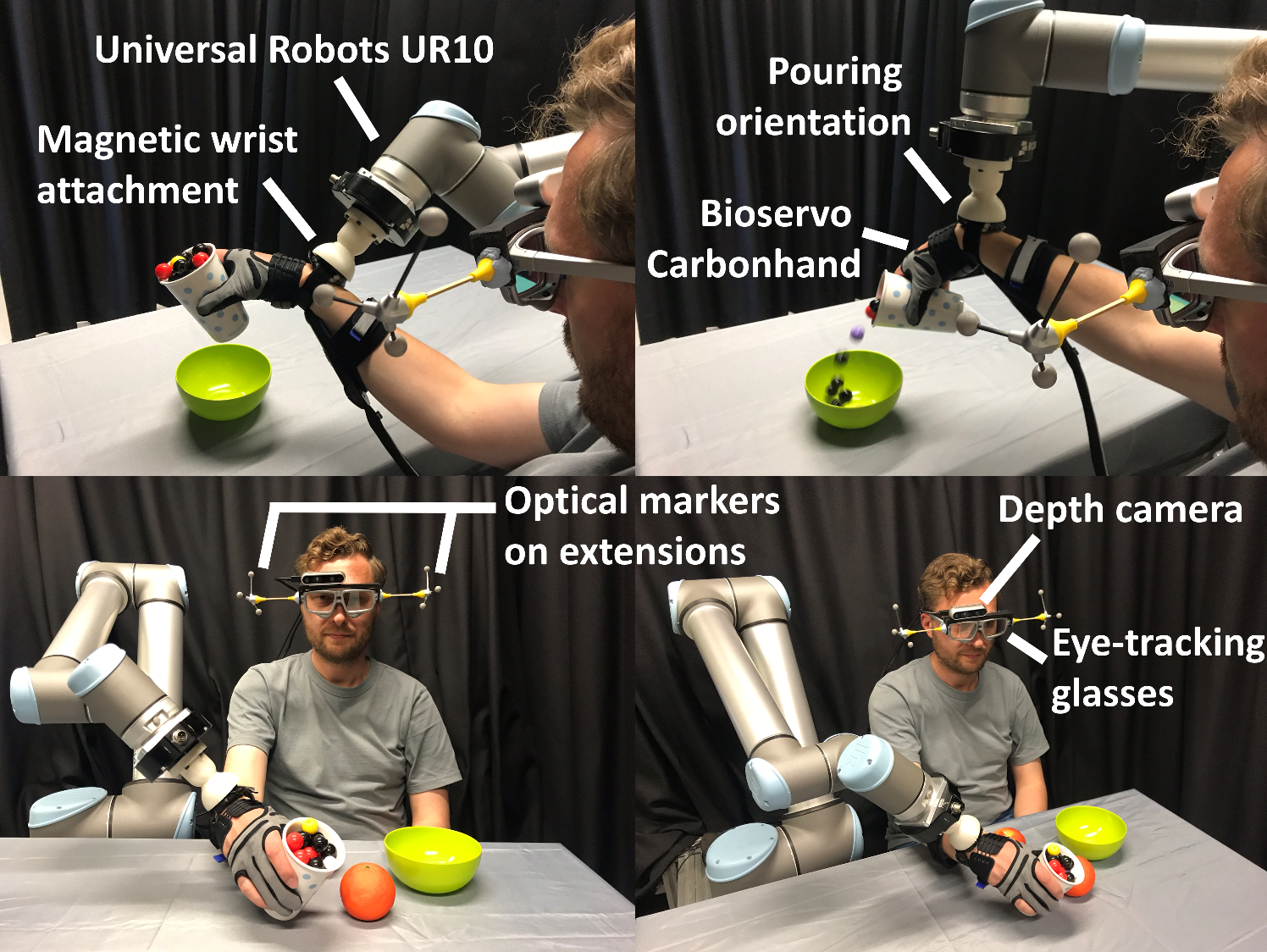}
\caption{A user with our robotic reach and grasp support system.}
\label{system}
\end{figure}
We have previously presented work on eye-tracking studies \cite{abbott2012ultra,tostado20163d} and the use of eye-tracking as a robot interface \cite{dziemian2016gaze}. In this work, we present a new method for 3D gaze point estimation, and its integration with a robotic system architecture allowing real-time intention decoding, decision making, and robot-actuated reach and grasp restoration. The system implements complex tasks by combining low-level actions, that are initiated with gaze, and carried through awareness of context and human action grammars.

Gaze has been used successfully in the past as an interface for machines, particularly in human-computer interfaces \cite{majaranta2014eye,olesova2017visual} and social robotics to monitor human attention, emotion and engagement \cite{breazeal1999context,yonezawa2007gaze,castellano2009detecting,he2018real} as well as robotic laparoscopic surgery \cite{fujii2018gaze}. When it comes to patients with movement disabilities, there is work on the use of gaze patterns in rehabilitation \cite{li2017eye}, for the control of 2 degrees of freedom in upper limb exoskeletons, where the patient uses gaze to direct the robot on a 2D surface. There is also work in assistive robotics in the form of tele-operated robots and wheelchairs controlled through gaze \cite{wang2018free,carlson2009using,gautam2014eye,6222082,bastos2014towards}. We aim to expand the research into 3D gaze monitoring for within assistive robotics for the \emph{restoration} of reaching and grasping. The use of gaze is of particular interest as it is retained in most upper limb disabilities. Furthermore, it allows for natural, easy to learn, and non-invasive interface between the human and the robot. To achieve this we rely on the idea of action grammars: our actions, like our sentences, have rules regarding how to combine them and which order to use to create a meaningful sequence.

In section II, we first present an overview of our system architecture followed by details of its consisting parts. Section III presents our evaluation experiments, with the results reported and discussed in Section IV. Section V concludes the paper, with proposals on future work.
%%%%%%%%%%%%%%%%%%%%%%%%%%%%%%%%%%%%%%%%%%%%%%%%%%%%%%%%%%%%%%%%%%%%%%%%%%%%%%%%
\section{Methods}
\subsection{System overview and architecture}
We aim to create a robotic system which acts based on human 3D gaze patterns and the context of the environment. Commercial eye-tracking glasses provide 2D gaze monitoring without any information on depth. We use an RGB-D camera mounted on top of the eye-tracker glasses to gain the missing depth information. We also need the user's head position and orientation, so that we can transform the 3D gaze points obtained within the eye-tracker's coordinate system, to that of the world. This information, along with the output of the object recognition module working on top of the eye-tracking ego-centric camera images, are then to be used to make decisions and implement actions using robotic devices.

\begin{figure}[tp]
\includegraphics[width=\columnwidth]{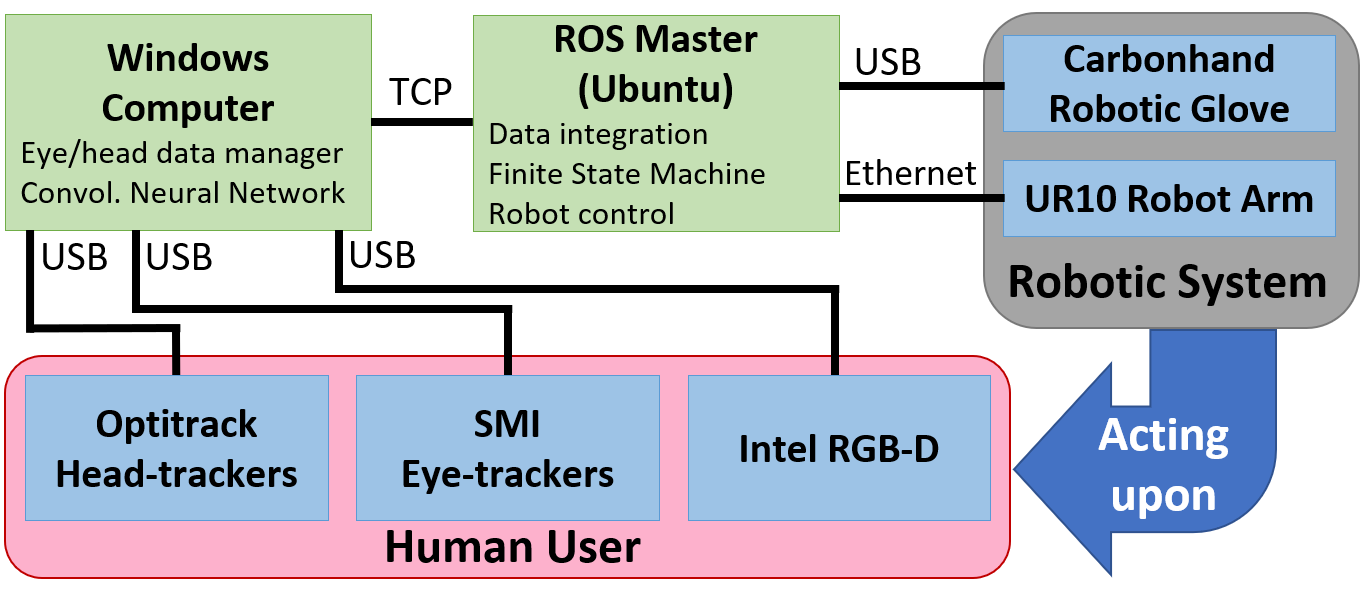}
\caption{Block diagram showing the architecture of our system.}
\label{architecture}
\end{figure}

Our system consists of multiple modalities integrated and working together through the Robot Operating System (ROS) environment \cite{quigley2009ros}. These include: 1. Eye-tracking glasses, 2. RGB-D camera, 3. Convolutional Neural Network for object recognition, 4. Optical head-tracking, 5. Robotic arm for reaching support and 6. Robotic glove for grasping support. The block diagram in Figure \ref{architecture} depicts an overview of our system. Individual modalities are described in detail in the following.
 \subsection{Eye-tracking}
 For 2D eye tracking, we use the SMI ETG 2W A (SensoMotoric Instruments Gesellschaft f\"{u}r innovative Sensorik mbH, Teltow, Germany). 
 %Using the SMI Software Development Kit (SDK), we are able to obtain 2D gaze positions superimposed on the ego-centric RGB camera image. To calculate 3D gaze points, we need depth information. 
 We mounted an Intel Realsense D435 RGB-D camera (Intel Corporation, Santa Clara, California, USA) on top of the eye-trackers, using a 3D printed frame. This can be seen in Figure \ref{system}. Typically, eye-tracking devices should be calibrated with the user's eyes. During calibration a user has to look at several points on a physical plane, and the researcher manually marks them on the ego-centric video feed (see Figure \ref{ettest}). In our setup, we use the depth picture from the RGB-D camera 
 %aligned to the camera's RGB image, as an ego-centric frame 
 and map gaze points directly to it during calibration. The Intel API for the depth camera provides the depth of each pixel in metres.
 
\begin{figure}[bp]
\includegraphics[width=\columnwidth]{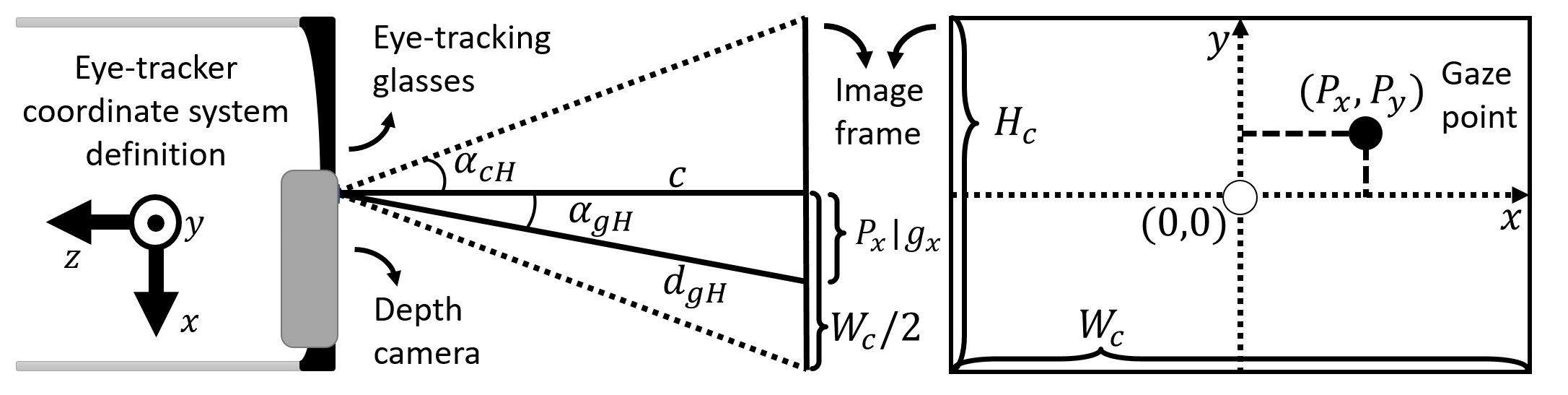}
\caption{Geometric representation of the gaze angle and the camera frame, used for the 3D gaze point estimation.}
\label{3dgp}
\end{figure}

 Through this integration, we are able to obtain the Euclidean distance, $d_g$, between the depth camera lens, and the surface in space over which the 3D gaze point of the user sits. The camera image resolution is $1280\times720$, let this be referred to as $W_c\times H_c$. The 2D gaze point is given in terms of pixels too, let this be represented as $(P_x,P_y)$, where $P_x$ is the horizontal gaze pixel location, and $P_y$ is the vertical one; with the centre of the image considered the origin, i.e. $(0,0)$ (see Figure \ref{3dgp}). We refer to the horizontal and vertical field of view (FOV) angles of the camera as $\alpha_{cH}$ and $\alpha_{cV}$, respectively. Let $c$ be defined as the line connecting the camera to the centre of its frame. The gaze angle is then defined in two cases. Horizontal gaze angle, $\alpha_{gH}$, is the angle between the projection of the Euclidean distance line above ($d_g$) on the horizontal (transverse) plane: $d_{gH}$. Similarly, the vertical gaze angle, $\alpha_{gV}$, is the angle between the projection of $d_g$ on the vertical (sagittal) plane: $d_{gV}$. The case of the horizontal gaze angle is displayed in Figure \ref{3dgp}.
We need to convert the gaze point values from pixels, to metres in Cartesian space: $(g_x,g_y,g_z)$. Consider the case shown in Figure \ref{3dgp}:
\begin{equation}
\begin{split}\label{eq:1}
     \tan(\alpha_{gH}) &= P_x/c \\
     \tan(\alpha_{cH}) &= (W_c/2)/c 
\end{split}    
\end{equation} 
Combining these we get:
\begin{equation} \label{eq:2}
    \alpha_{gH} = \arctan(\frac{P_x}{W_c/2}\tan(\alpha_cH)) 
\end{equation}
We now know the angle between the Euclidian distance gaze line, $d_g$, and the centre line of the camera frame. We obtain $\alpha_{gV}$ similarly. We know that:
\begin{equation} \label{eq:3}
\begin{split}
     g_x = d_{gH}\sin(\alpha_{gH}) \\
     g_y = d_{gV}\sin(\alpha_{gH})
\end{split}
\end{equation}
Also, considering the right-angled triangles formed between $d_g$ and its projections, we can write:
\begin{equation} \label{eq:4}
\begin{split}
     g_x^2 + d_{gV}^2 = d_g^2 \\
     g_y^2 + d_{gH}^2 = d_g^2
\end{split}
\end{equation}
Combining equations \ref{eq:3} and \ref{eq:4} yields a system of linear equations as follows:
\begin{equation} \label{eq:5}
\begin{cases}
\sin^2(\alpha_{gH})d_{gH}^2 + d_{gV}^2 = d_g^2 \\
d_{gH}^2 + \sin^2(\alpha_{gV})d_{gV}^2 = d_g^2 
\end{cases}
\end{equation}
which we can solve to find $d_{gH}$ and $d_{gV}$, and use them to obtain $g_x$ and $g_y$ from equation \ref{eq:3}, and from there, find $g_z$.
%Then, to obtain the z-distance of the gaze point:
% \begin{equation} \label{eq:6}
%     g_z = \sqrt{d_g^2 - g_x^2 - g_y^2}
% \end{equation}

We now have the 3D gaze point of the user, in Cartesian coordinates, with respect to the defined eye-tracker coordinate system. We then need to transform this to the robot coordinate system. To do this, we need the position and orientation of the user's head.
\subsection{Optical head tracking}
We use the Optitrack Flex 13 cameras (NaturalPoint, Inc. DBA OptiTrack, Corvallis, Oregon, USA) for optical head tracking. To avoid occlusions due to the user's head, we created 3D printed extensions. This can be seen in Figure \ref{system}. 
% The markers are then selected in the Optitrack software, Motive; and a rigid body is defined. 
% This method gives us the position of the centre of the rigid body (i.e. the camera depth lens position) with respect to the pre-defined Optitrack origin, and the rotation of the rigid body, relative to its initial position when selected and defined in Motive. We use these values to form a transformation matrix which is applied to the 3D gaze point coordinates transforming them into the Optitrack coordinate system followed by an extra transformation matrix to transform this into the robot coordinate system.
This method gives us the pose of the camera with respect to the pre-defined Optitrack origin. We use these to transform 3D gaze position into the Optitrack coordinate system, and from there to the robot coordinate system.
\subsection{Detection of objects and intention of action}
We use a deep neural network approach coupled with naive classification to classify multiple objects in the user's field of view. The development of this system is highlighted in \cite{auepanwiriyakul2018semantic}. The output of this is real-time object recognition on the depth camera images, with rectangular bounding boxes drawn around the detected object. We can then use the detected gaze position on the camera image frame, along with these bounding boxes to detect: 1. which object and 2. which part of that object, is the user gazing upon. We use this information to extract context and intention - i.e. which objects is the user interested in, and whether there is an intention of physical interaction with this object. To detect the latter, we have defined the right-hand side of each object, as the location for the user to gaze at (for 15 gaze points), to indicate an intention of physical motion. 
% This is implemented based on the bounding boxes provided by the object recognition module -- we divide the box into three equal boxes horizontally, and consider the right-most box as the one to indicate intention with. This is applicable to complex and simple objects alike, as the algorithm always draws rectangular bounding boxes. 
This approach gives the user executive control, allowing them to freely inspect objects without causing robot movements. 
\subsection{Robotic system integration}
Our robotic system consists of two commercial robots: The Universal Robots UR10 (Universal Robots A/S, Odense, Denmark) and the BioServo Carbonhand (Bioservo Technologies AB, Kista, Sweden). The former is used for reaching and the latter for grasping support. The user wears the robotic glove on their hand, and attaches their arm to the UR10 through a 3D printed magnetic attachment on their wrist. The magnetic setup is used to ensure our test subjects are able to detach their arm by pulling it away if they sense a risk. Strict workspaces, motion planning constraints including a 3D reconstruction of our lab environment and user bounding boxes for collision avoidance are in place to ensure safety. As we use ROS, the robot choice is irrelevant, as long as it is ROS-compatible.

% The ROS master receives the gaze point as pixel locations within the camera frame, the object that the user's gaze falls upon and whether there is an intention of motion as well as the user's head position and orientation. 
The 3D gaze point calculations described above are implemented within our ROS package. We then have the user's 3D gaze point, i.e. the location of the object they are looking at, as well as knowledge on what that object is and whether the user wants to physically interact with that object. We use these inputs to make decisions and implement sequences of actions with the robotic system, using low-level actions and following rules of action grammars. A finite state machine (FSM) is applied to implement this.
%The intention of motion above serves as a trigger for the robotic system, acting as an implicit command by the user. However, we want the entire sequence of actions that constitute the task at hand to be performed upon receipt of that trigger. For this, we rely on the idea of action grammars highlighted in the introduction, i.e. the context will limit what actions are possible to follow in the sequence. This allows us to implement tasks as a series of lower level actions with rules on how they form sequences. To implement this, as a cognitive modelling of human action behaviour, we use a finite state machine (FSM).
%%%%
\begin{table}[tp]
    \captionsetup{justification=centering}
    \caption{Convention used for the categories of objects within the finite state machine.}
    \label{tab:objects}
    \centering
    \begin{tabular}{|c|c|c|c|}
    \hline
    Graspable (G) & Pourable (P) & GP & Comments
\\ \hline 
    0 & 0 & 00 & e.g. large container
\\ \hline 
    0 & 1 & 01 & undefined e.g. table
\\ \hline 
    1 & 0 & 10 & e.g. apples/oranges
\\ \hline 
    1 & 1 & 11 &  e.g. small container
\\ \hline
    \end{tabular}
% \vspace{-8mm}
\end{table}
%%%%

As an example for our proof of concept, we are dealing with a dining table scenario. We are therefore looking at objects such as fruits (apples, oranges) and containers (cups, bottles, bowls). We define the interaction between these objects as 1. pick and place on the table, 2. pick and place into containers and 3. pick and pour into larger containers. Grammars are already visible here, i.e. you can pick and place fruits on the table or into the bowl, but not into the cup or bottle; similarly, you cannot pour fruits, but you can pour the cup/bottle -- and only into the bowl and not on the table. We categorise our scenario objects considering their Graspability ($G$) and Pourability ($P$) as defining parameters. For example, apples and oranges are graspable but not pourable, cups are graspable and pourable. We use a binary string for notation of objects. See Table \ref{tab:objects} for a description. Note that a non-graspable but pourable object is undefined - this category, i.e. $GP=01$ is used to represent the dining table itself.
%%%%
\begin{table}[bp]
    \captionsetup{justification=centering}
    \caption{Convention used to name the states within the finite state machine.}
    \label{tab:states}
    \centering
    \begin{tabular}{|c|c|c|}
    \hline
    Grip & Object held & Comments
\\ \hline 
    0 & 01 & 001: Grip open, no object held.
\\ \hline 
    1 & 01 & 101: Grip closed, no object (grasp failure)
\\ \hline 
    1 & 10 & 110: Grip closed, graspable non-pourable
\\ \hline 
    1 & 11 & 111: Grip closed, graspable pourable
\\ \hline
    \end{tabular}
% \vspace{-8mm}
\end{table}
%%%%

The states of the FSM are defined to represent the user's states. The parameters used in this definition are whether the user's grip is open or closed, and what object is held in their grip, if any. We represent this in a binary string format too, with the grip open being represented as $0$ and grip closed as $1$. This is followed by the object held, using the same notation as that of Table \ref{tab:objects}, except that in this case, $GP=01$ is used to represent "no object held". This leads to a total of 4 states for the FSM, which are listed in Table \ref{tab:states}. 
%Note that there is no state relating to object held being of type $GP=00$ (non-graspable, non-pourable) - as these are non-graspable, such as the large bowl. Similarly, with the grip open, there is only one type of object held: none, as obviously the user cannot be holding an object with their grip open.

\begin{figure}[tp]
\includegraphics[width=\columnwidth]{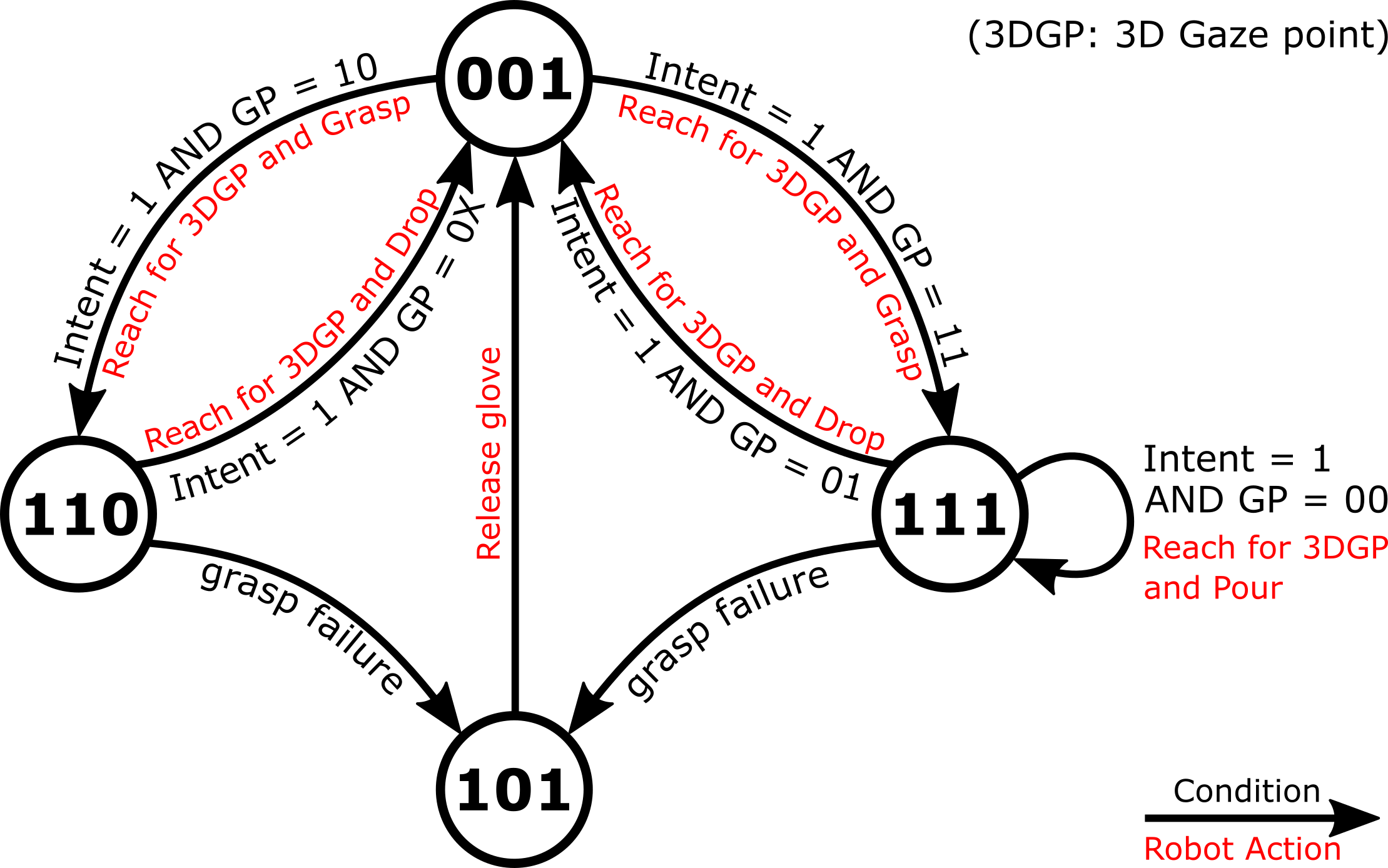}
\caption{The finite state machine used to implement the sequences of action.}
\label{fsm}
\end{figure}

The full FSM, with the states, transitions, their conditions and actions, is depicted in Figure \ref{fsm}. The black text on each transition arrow defines the conditions for that particular transition. Note that the intention of action as detected from the user's gaze patterns, is represented as a Boolean variable here named “Intent”. The $GP$ condition on the transition arrows relate to the object gazed by the user, following the convention of Table \ref{tab:objects}. The red text on each transition arrow indicates the robotic action that is triggered by that transition.
% , i.e. “Reach for 3DGP and Grasp” means the UR10 robotic arm will reach for the 3D gaze point of the user, and once this is completed, the Carbonhand robotic glove will close the user's grip.

The user starts in state $001$ (grip open, no object held) and remains in this state as long as no intention of action is detected. Note that self-transitions in case of unfulfilled conditions are not displayed in the FSM figure for simplicity. Once the user's gaze pattern indicates an intention of action ($intent == 1$), depending on the object the user is looking at, one of the following will occur: Looking at a Graspable, non-Pourable object ($GP=10$, e.g. apple, orange), the machine will transition to $110$. The robotic system will reach for the object and grasp - and similarly for looking at $GP=11$ (e.g. cup, bottle) it will transition to $111$. If the user is looking at a non-Graspable, non-Pourable object ($GP=00$, e.g. bowl) or the table ($GP=01$), the machine will not transition. Note that transitions will not execute if the 3D gaze point of the user is not within the workspace, or if it is not motion planable for the robot.
%%
% If the machine has transitioned into state $111$, this means the user is now holding a graspable, pourable object. The possible actions are either to hold it, pour it into a container or drop it on the table. Holding it would mean remaining in the same state – which will be the case as long as there is no intention of action. Once there is an intention of action, and the user looks at $GP=00$, i.e. at a container, the transition will be executed: the robot will move to the 3D gaze point and perform a pouring action, rotating the user's wrist over the container. For the sake of this simulated scenario, the pouring action is timed, i.e. the robot will rotate the user's wrist into a pouring orientation, hold it for 3 seconds, and then rotate it back into a holding orientation. When this transition and action are complete, the user is still holding a pourable object, therefore the state remains $111$ – hence the self transitioning arrow in Figure \ref{fsm}.

% In state $111$, if there is an intention of action and the user looks at $GP=01$, i.e. at the table, at a point within the safe workspace and reachable by the robot; then the transition will be executed: the robot will move to the 3D gaze point and drop the object there. At this point the user is not holding any object and their grip is open – hence the state is $001$.
The `grasp failure' transition is to handle the potential cases when the glove closes but the grasp is not successful, or the case when a grasp has been made, but it is not stable and the object is dropped midway through the task. 
% Failure of a grasp can only initiate from a state which involves an object having been grasped already – i.e. $110$ or $111$. From these two states, if the grasp is unsuccessful, a transition will be made to state $101$. Following the convention of Table \ref{tab:states}, $101$ means that the grip is closed but no object is held. 
Such cases lead to state $101$ (grip closed, no object held).
This state will immediately transition to $001$ by releasing the grip. The ROS package for the Carbonhand robotic glove publishes tendon tension values, motor voltages and force sensor values from the glove finger tips. Combining these data, we are able to detect 1. whether the glove is closed or open and 2. if closed, whether the user is holding an object, or an empty grip. This is used to detect the user's state and particularly grasp failures.

Note that in practical implementation, there is a clear offset from the user's grasp point, to the robot TCP, which depends on the size and orientation of the magnetic attachment, as well as each user's particular wrist diameter, hand size and finger length (Figure \ref{system}). To personalise the system to each user, we created a calibration step: We 
% move the robot with the user's arm attached to it to an arbitrary point on the table, at a comfortable grasp height. We place a cup (can be any object) within their grasp reach and ask them to gaze upon it. The system records the calculated 3D gaze point, and the real-time robot position, subtracting the two to find the offset in all three axis. This is then stored and used throughout trials for that user. 
ask the users to look at the palm of their hand, while attached to the robot. We use the calculated 3D gaze point and the real-time robot position to calculate this offset, which is stored and used throughout the trial.

We have a fully functioning FSM that once activated can lead to continuous action implementations by the user without any interference by the system technicians. The grasp failure state allows for even failed tasks to simply be repeated until successful. 
%As can be seen from Figure \ref{system}, the magnetic attachment holds the user's grip at an angle. This is to allow for the user to remove their arm in case of unexpected movements, and to not have them trapped under the robot (i.e. as opposed to the case of the robot TCP facing the floor). In this orientation, to pour, the robot changes the orientation of the TCP to face down, which will lead to the cup being held parallel to the floor and pouring successfully (Figure \ref{system}). This change in orientation is accompanied by a relevant change in position to keep the position of the tip of the cup constant. Furthermore, there is a clear offset from the user's grasp point, to the robot TCP, which depends on the size and orientation of the magnetic attachment, as well as each user's particular wrist diameter, hand size and finger length. To personalise the system to each user, we have added a calibration functionality as follows: We move the robot with the user's arm attached to it to an arbitrary point on the table, at a comfortable grasp height. We then place a cup (can be any object) within their grasp reach and ask them to gaze upon it. The system records the calculated 3D gaze point, and the real-time robot position, subtracting the two to find the offset in all three axis. This is then stored and used throughout trials for that user. 
%%%%%%%%%%%%%%%%%%%%%%%%%%%%%%%%%%%%%%%%%%%%%%%%%%%%%%%%%%%%%%%%%%%%%%%%%%%%%%%%
\section{Experimental Evaluation}
\subsection{Evalutation of the 3D gaze calculation method}
\begin{figure}[tp]
\includegraphics[width=\columnwidth]{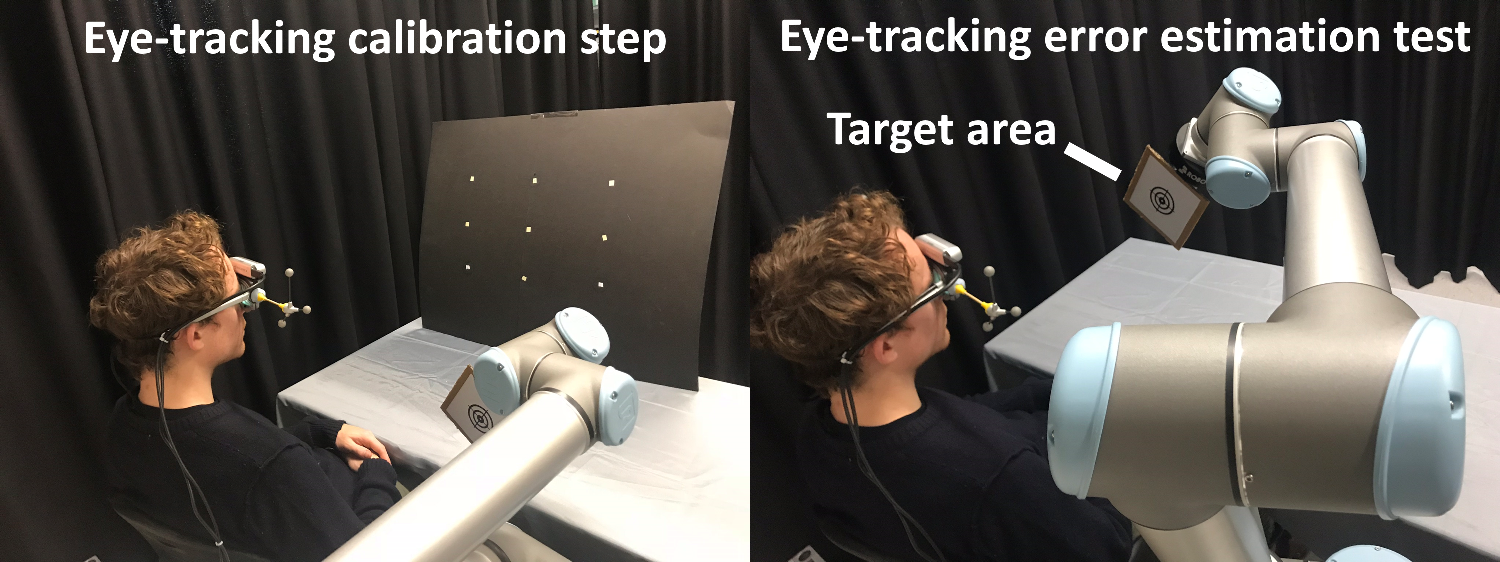}
\caption{The eye-tracking evaluation test setup.}
\label{ettest}
\end{figure}
To evaluate the accuracy of the 3D gaze estimation method, we programmed the UR10 robot to move to 10 points obtained randomly with uniform distribution, within the following range: $([0.25,0.80],[0.15,0.75],[0.35,0.75])$, with a target attached at its end-effector. At each point, 50 gaze samples are obtained along with the real-time robot position. The eye-trackers are running at $60Hz$, but the real-time object recognition algorithm slows down the pipeline. The final calculated 3D gaze values are published at a frequency of $\approx10Hz$, this is therefore equivalent to about 5 seconds at each point. The SMI eye-trackers require an initial period of random eye movements to obtain the pupil positions followed by a 3-point calibration.
%where the user fixates on three points with the system operator clicking on the screen at the point of fixation. 
The full experiment setup is shown in Figure \ref{ettest}. The 9-point board is placed at an $80cm$ distance from the user. Each subject is first asked to randomly fixate on the points for 1 minute. This is followed by the calibration step which is performed on the bottom left, top centre, and bottom right points on the board. The calibration is then verified and the board removed before the actual experiment starts. Errors in gaze estimation in x-axis, y-axis, z-axis and as Euclidean distance are measured and recorded. Results of the evaluation are provided and discussed in section \ref{sec: results}.
\subsection{Evaluation of the integrated robotic system}
\begin{figure}[bp]
\includegraphics[width=\columnwidth]{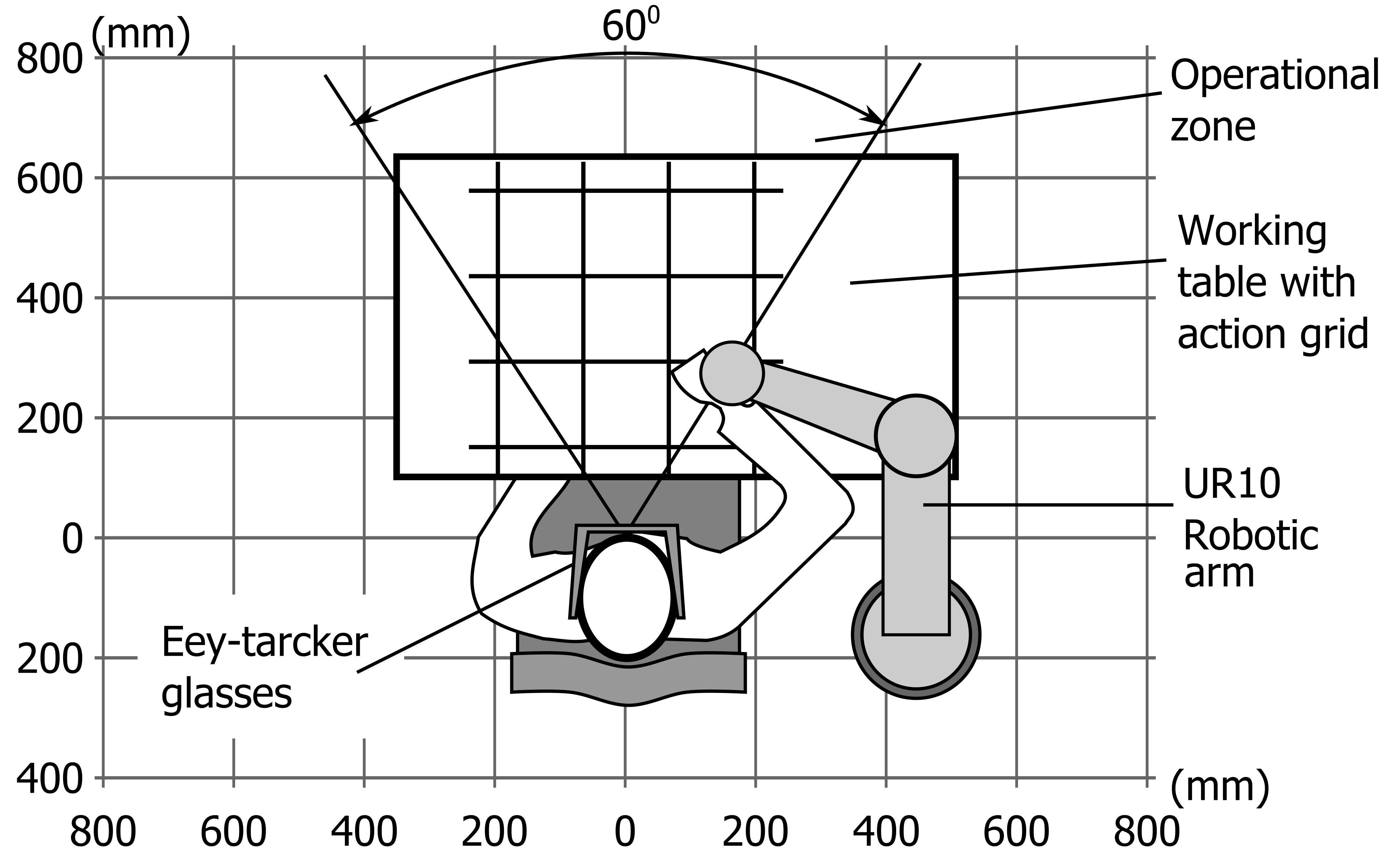}
\caption{The full integrated robotic system test setup.}
\label{fulltest}
\end{figure}
To evaluate the integrated system in an efficient manner, we selected the cup, bowl and table as objects of interaction. This is due to the cup having both "pick and place" and "pick and pour" functionalities within it, showing the grammar and context-based approach implemented with the FSM. The system functions similarly with other objects discussed before, e.g. apples, oranges - these are shown in our video attachment.

A grid is drawn on the table in front of the user, creating 9 square boxes. The placement of the cup and bowl, and the target point on the table to drop objects on is randomised between these boxes for each trial. We chose to use the grid setup so that we would have a measure of success for object placement on the table, without indicating an exact point to our participants that they can fixate on, as doing so might help with the eye-tracking accuracy. The boxes are of an approximate size of $13cm\times13cm$; but this is not an indication of the system resolution, which is instead reported in section \ref{sec: results}. A schematic of the experiment setup can be seen in Figure \ref{fulltest}. 

The experiment tasks are: 1. Pick up the cup and Place it back on the table at a different location (PP) and 2. Pick up the cup, Pour it into the bowl, and then Place it at a different location on the table (PPP). For pouring, small plastic balls are used to simulate a liquid, while conserving health and safety. Note that throughout the experiments, one of the researchers is constantly in possession of the UR10's emergency stop button, for added safety. Tasks are to be performed 5 times each. The users are asked not to contribute to the actuation and allow actions to be performed by the robotic arm and glove. As there is a learning curve involved with using the system, 3 attempts were allowed for each trial's first reach action; failures at later stages of a task are considered a task failure. Tasks are broken down into their low-level actions, and the success/failure of these as well as the overall task success/failure are recorded as outcomes. Each participant completes a System Usability Scale (SUS) \cite{brooke1996sus} subjective questionnaire after their test. These results are presented and discussed in section \ref{sec: results}.

%%%%%%%%%%%%%%%%%%%%%%%%%%%%%%%%%%%%%%%%%%%%%%%%%%%%%%%%%%%%%%%%%%%%%%%%%%%%%%%%
\section{Results and Discussion} \label{sec: results}
\subsection{Gaze estimation results}
For the evaluation of 3D gaze estimation, 8 subjects were invited to our study, age range of $27.5\pm2.5$, 6 male and 2 female. All subjects had normal or corrected to normal vision (wearing glasses which they were asked to remove). The first 10 gaze points from each trial are filtered out to avoid the transient effect of the user's gaze. We use the 40 remaining gaze points for analysis, a total of 3280 gaze points. Each point has 3D coordinates of gaze location and ground truth (i.e. robot position). We average 3D coordinates of gaze data per trial to filter the gaze noise resulting in 80 data points. We use the Euclidean distance between the calculated gaze point and ground truth as the measure of accuracy.

\begin{figure}[tp]
\includegraphics[width=\columnwidth]{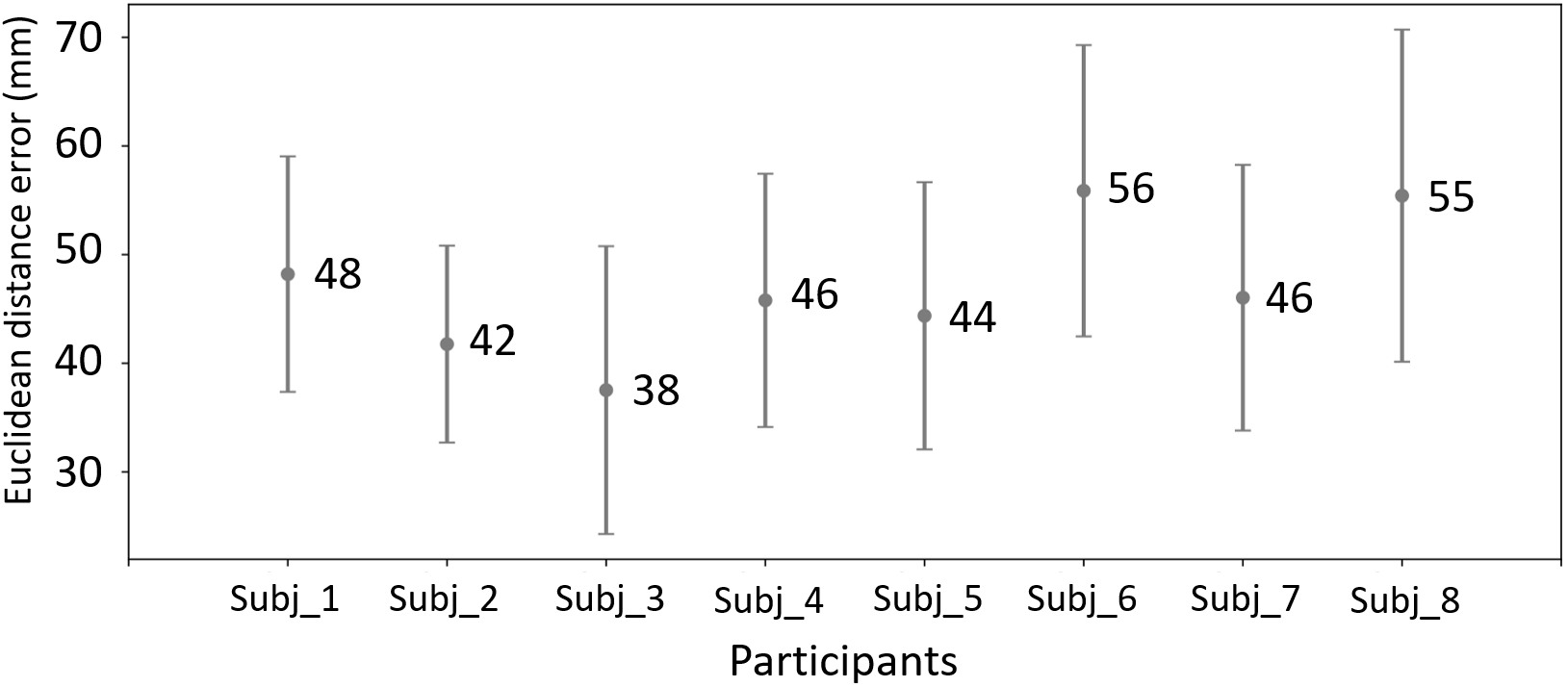}
\caption{The eye-tracking evaluation results: Euclidean distance error in millimetres for all 8 participants as mean and standard deviation.}
\label{result1}
\end{figure}

In average, our system performs with the Euclidean error distance of $4.68\pm0.014cm$ (mean\textpm SD). The Euclidean distance is normally distributed with $0.001$ level (D'Agostino and Pearson's normality test: $p=0.029$). To inspect the possible human factor influence, we perform one-way ANOVA. We found that the human factor does not affect Euclidean distance significantly with $0.001$ level ($F(7,72)=2.345,p=0.032$). Figure \ref{result1} shows the mean and standard deviation of the measure per subject. We can conclude that all subjects perform similarly.
%%
%%
% \begin{figure}[bp]
% \includegraphics[width=\columnwidth]{result2.png}
% \caption{The eye-tracking evaluation test results with respect to the three axes of measurement.}
% \label{result2}
% \end{figure}
%%

We also checked the accuracy with respect to individual axes. We did not find significant correlation for x-axis (Spearman rank: $p=0.085$), or the y-axis errors (Spearman rank: $p=0.715$), which correspond with the user's depth and horizontal axes directions respectively. The Z-axis or the height from the user's perspective, has significant correlation (Spearman rank: $correlation=-0.460; p << 0.01$). We believe this is due to the lower accuracy of pupil-tracking in extreme vertical positions. When eye balls go up they are less visible for the infrared cameras of the eye-tracker glasses. Overall, the results are consistent and show good performance of the system. 

\subsection{Full system results}

For our full system tests, 5 participants joined the study. These were all male, $30.0\pm5.0$ of age. Each experiment task consists of a number of low-level actions. For pick and place on table (PP in Table \ref{tab:results}), the actions are: Reach1, Grasp, Reach2 and Drop. For pick, pour in bowl, place on table (PPP in Table \ref{tab:results}), the actions are: Reach1, Grasp, Reach2, Pour, Reach3, Drop.
%%%
\begin{table}[bp]
\captionsetup{justification=centering}
\caption{Success percentages per action, per task, per participant, when using the full robot integrated system. PPP is "Pick, Pour and Place" and PP is "Pick and Place". A total of 5 participants perform the two tasks, repeating each of them 5 times. We report percentage of successful actions within those repeats, for each participant.}
\label{tab:results}
\resizebox{\columnwidth}{!}{%
\begin{tabular}{|c|c|c|c|c|c|ll}
\hline
\textbf{ID} & \textbf{Task} & \textbf{Reach1} & \textbf{Grasp} & \textbf{Reach2} & \textbf{Pour} & \multicolumn{1}{c|}{\textbf{Reach3}} & \multicolumn{1}{c|}{\textbf{Drop}} \\ \hline
P1 & \multirow{5}{*}{\textbf{PPP}} & 100\% & 100\% & 100\% & 80\% & \multicolumn{1}{c|}{100\%} & \multicolumn{1}{c|}{50\%} \\ \cline{1-1} \cline{3-8} 
P2 &  & 100\% & 100\% & 100\% & 100\% & \multicolumn{1}{c|}{100\%} & \multicolumn{1}{c|}{80\%} \\ \cline{1-1} \cline{3-8} 
P3 &  & 100\% & 100\% & 100\% & 80\% & \multicolumn{1}{c|}{100\%} & \multicolumn{1}{c|}{100\%} \\ \cline{1-1} \cline{3-8} 
P4 &  & 100\% & 100\% & 100\% & 100\% & \multicolumn{1}{c|}{100\%} & \multicolumn{1}{c|}{100\%} \\ \cline{1-1} \cline{3-8} 
P5 &  & 100\% & 80\% & 100\% & 100\% & \multicolumn{1}{c|}{100\%} & \multicolumn{1}{c|}{100\%} \\ \hline
\textbf{ID} & \textbf{Task} & \textbf{Reach1} & \textbf{Grasp} & \textbf{Reach2} & \textbf{Drop} &  &  \\ \cline{1-6}
P1 & \multirow{5}{*}{\textbf{PP}} & 100\% & 80\% & 100\% & 100\% &  &  \\ \cline{1-1} \cline{3-6}
P2 &  & 100\% & 100\% & 100\% & 100\% &  &  \\ \cline{1-1} \cline{3-6}
P3 &  & 100\% & 100\% & 100\% & 100\% &  &  \\ \cline{1-1} \cline{3-6}
P4 &  & 100\% & 100\% & 100\% & 100\% &  &  \\ \cline{1-1} \cline{3-6}
P5 &  & 100\% & 100\% & 100\% & 80\% &  &  \\ \cline{1-6}
\end{tabular}%
}
\end{table}
%%%
Overall, for all participants and across all trials, PP was performed successfully in full $96\%$ of the time. There were only two cases of failure: 1 instance of failing to grasp the cup and 1 instance of dropping the object slightly outside the indicated target box. Therefore, for PP, across all trials and participants, Reach1 was successful $100\%$ of the time, Grasp $96\%$, Reach2 $100\%$ and Drop $96\%$ of the time. PPP task was performed successfully in full $76\%$ of the time. Failures were 3 instances of the final drop (was in the target area, but not placed upright); 2 instances of pour failures (pouring orientation change led to dropping the cup) and 1 instance of failure in the initial grasp of the cup. Therefore, for PPP, across all trials and participants, Reach1 was successful $100\%$ of the time, Grasp $96\%$, Reach2 $100\%$, Pour $91.7\%$, Reach3 $100\%$ and the final Drop, $87\%$ of the time. 

These results show mainly that the 3D gaze point estimation is well integrated with the system: all reaching cases are $100\%$ successful, which is fully dependent on the 3D gaze point estimation being accurate. The Finite State Machine performed without errors throughout, making the implementation of these complex tasks possible with a very short training period (less than 5 minutes in all participants). Observed issues in the results are mainly within the pouring task, particularly at the pour action and its aftermath. The pouring orientation change had the effect of slightly moving the cup within the subjects' grasp (the cup and its contents are heavy), leading to either a premature drop of the cup, or a badly placed drop later on. This is mainly due to the design of the magnetic wrist attachment. We realised throughout the experiments that it does not provide robust support for the pouring action. This is an item that can be improved in the future.

All 5 participants filled in the System Usability Scale after their tests, the results are summarised in table \ref{tab:sus}.
%Two questions on the original SUS which were deemed not applicable, were removed. These are: 1. “I think that I would like to use this system frequently” – the end-users of this system are people with reach/grasp disabilities, it would therefore be non-applicable for healthy participants to consider whether they would use it frequently: a healthy person would not use this system. 2. “I think that I would need assistance from technicians to be able to use this system.” – this is more applicable for a product, not a technical prototype in a lab environment being run be researchers – obviously all participants needed assistance to use the system in its current form. Opinions on the system being "unnecessarily complex" are divided - 3 out of 5 choosing the borderline option, 1 agreed and 1 disagreed. On system "ease of use", 3 agree, 1 borderline and 1 disagree. On the system being "well integrated", 4 out of 5 agree, and 1 is borderline. On the system being "unpredictable" opinions are very divided: 1 agreed, 2 borderline, 1 disagreed and 1 strongly disagreed. On whether "most people would learn to use the system quickly", 4 out of 5 agreed (2 strong agreements) and 1 is borderline. On whether the system is "cumbersome to use", 3 disagreed and 2 are borderline. On whether they felt "confident using the system", 2 agree (1 strongly), 2 are borderline and 1 disagrees. On whether they "needed to learn a lot before they could use the system", all users disagree - 2 of them strongly.
These results show that, for our participants, the system was easy to learn and not cumbersome. Most division of opinions are on whether the system is unpredictable and whether they felt confident using the system -- though even in these cases results are favouring the system. We believe these two issues are related. As the users receive no direct feedback on how and when decisions and actions are made, the behaviour of the system might seem unpredictable, which will result in the users feeling less confident in its behaviour. This is an issue for us to look into as future work.
%%%
\begin{table}[tp]
\captionsetup{justification=centering}
\caption{Results of the System Usability Scale \cite{brooke1996sus} questionnaire as filled out by the five participants. Each row lists a question, and how many participants chose which answer.}
\label{tab:sus}
\resizebox{\columnwidth}{!}{%
\begin{tabular}{|c|c|c|c|c|c|}
\hline
\textbf{Question about system} & \textbf{\begin{tabular}[c]{@{}c@{}}Strongly \\ Disagree\end{tabular}} & \textbf{Disagree} & \textbf{Borderline} & \textbf{Agree} & \textbf{\begin{tabular}[c]{@{}c@{}}Strongly \\ Agree\end{tabular}} \\ \hline
\textbf{unnecessarily complex?} & 0 & 1 & 3 & 1 & 0 \\ \hline
\textbf{easy to use?} & 0 & 1 & 1 & 2 & 1 \\ \hline
\textbf{well integrated?} & 0 & 0 & 1 & 4 & 0 \\ \hline
\textbf{unpredictable behaviour?} & 1 & 1 & 2 & 1 & 0 \\ \hline
\textbf{can be learned quickly?} & 0 & 0 & 1 & 2 & 2 \\ \hline
\textbf{very cumbersome to use?} & 0 & 3 & 2 & 0 & 0 \\ \hline
\textbf{felt confident using it?} & 0 & 1 & 2 & 1 & 1 \\ \hline
\textbf{a lot to learn before using it?} & 2 & 3 & 0 & 0 & 0 \\ \hline
\end{tabular}%
}
\end{table}
%%%
%%%%%%%%%%%%%%%%%%%%%%%%%%%%%%%%%%%%%%%%%%%%%%%%%%%%%%%%%%%%%%%%%%%%%%%%%%%%%%%%
\section{Conclusions}
We presented a gaze-contingent robotic system for the restoration of reach and grasp capabilities. The main focus of our approach was to create a non-invasive, easy-to-learn and easy-to-use interface that would allow implementation of complex tasks made of sequences of several low-level actions while conserving the simplicity of the interface. We follow the idea of action grammars: there are rules to our actions and how they can be combined together to create complex tasks. We implemented this, as a modelling of human cognitive behaviour, in the form of a finite state machine which monitors the human state, and implements actions on the robotic system when the necessary conditions are in place. This adds safety to the interaction, while making the implementation of complex tasks easy. 

The user's gaze is monitored in high stability and accuracy through a new 3D gaze estimation method, by integrating eye-tracker glasses and a depth camera. The objects within the user's environment are recognised using a convolutional neural network running on ego-centric camera images, allowing us to understand the context of the user's environment and interaction. Combining these, we are able to understand which object the user is looking at, whether they are interested in interacting with that object, and where that object is located in 3D space. 

With the human state and environment context fed to the finite state machine, it can now direct the robotic system to implement complex sequences of actions for the user. Example tasks of "pick and place", as well as "pick, pour and place" were run with participants to test the system's performance and usabiltiy. Note that these are examples and that actions and tasks can be expanded without issues; the finite state machine enquires a database of actions and grammars to fulfil these tasks which can simply be extended with further actions and grammars. The same applies to the object detection method -- it can be extended with different datasets. Results showed successful implementation of $100\%$ of reaching actions, as well as $96\%$ success in the "pick and place" task, and $76\%$ in the "pick, pour and place" task. Issues in task completion were not related to the performance of the 3D gaze estimation, object recognition or finite state machine modules; but rather due to physical and mechanical design choices within the system, such as the magnetic wrist attachment which is not the best support for pouring actions. Since then we have developed new, improved wrist attachments -- an example of this is shown in our video attachment.

The participants also filled in subjective questionnaires which showed they were content with ease of use and integration of the system, but were not all feeling confident with the system, with some feeling that it is unpredictable. We believe this is due to lack of feedback to the users indicating an imminent action, which results in lack of explainability and therefore unpredictability. Furthermore, due to the limitation in experiment population (N=5) and demographic (healthy, young, male) the results need to be considered with care. We will run more experiments with larger populations and more diverse demographics to remedy this. We will also look into improved feedback to the user, and general user experience as future work. We are interested in making further use of the RGB-D camera, particularly to implement SLAM for better understanding of the environment (e.g. for obstacle avoidance when moving around the table), and better tracking of the user's head, to possibly remove the Optitrack system entirely making the system more mobile, e.g. through the use of a wheelchair mounted robotic arm for reaching support. These will be pursued as future work. 

Aside from the particular use-case described here, we present this work as a Human-Robot Interaction interface. The approach can be used for any case of HRI which requires understanding of human visual attention and intention, leading to implementation of sequences of actions. This can be useful for assistive robots, humanoids and social robots, autonomous cars, or collaborative robots working with humans on a joint task, to name a few.


\bibliographystyle{ieeetr}
\bibliography{refs.bib}

\end{document}